\documentclass[sigconf,screen,nonacm]{acmart}
\usepackage{multirow}
\usepackage{hyperref}
\usepackage{xcolor}
\usepackage{xspace}
\usepackage{xurl}
\usepackage{tabularx}

\newcommand{\VEE}{VEE\xspace}
\newcommand{\VEC}{VEC\xspace}
\newcommand{\VECs}{VECs\xspace}
\newcommand{\VAEER}{VAEER\xspace}
\newcommand{\VAEERL}{VAEER$_L$\xspace}
\newcommand{\VAEERJ}{VAEER$_J$\xspace}
\AtBeginDocument{%
  }

\newenvironment{packed_itemize}{
\vspace{-\topsep}
\begin{list}{\labelitemi}{\leftmargin=1.5em}
  \setlength{\itemsep}{1pt}
  \setlength{\parskip}{0pt}
  \setlength{\parsep}{0pt}
  \setlength{\headsep}{0pt}
  \setlength{\topskip}{0pt}
  \setlength{\topmargin}{0pt}
  \setlength{\topsep}{0pt}
  \setlength{\partopsep}{0pt}
}{\end{list}\vspace{-\topsep}}

\setcopyright{acmlicensed}
\copyrightyear{2025}
\acmYear{2025}
\acmDOI{XXXXXXX.XXXXXXX}
\acmConference[WWW '26]{The Web Conference 2026}{April 13-17, 2026}{Dubai, United Arab Emirates}
\acmISBN{978-1-4503-XXXX-X/2025/10}

\begin{document}

\title{\VAEER: Visual Attention-Inspired Emotion Elicitation Reasoning}


\author{Fanhang Man}
\orcid{0000-0002-5830-0685}
\affiliation{%
  \institution{Tsinghua Shenzhen International Graduate School, Tsinghua University}
  \city{Shenzhen}
  \country{China}
}
\email{mfh21@mails.tsinghua.edu.cn}

\author{Xiaoyue Chen}
\affiliation{%
  \institution{Tsinghua Shenzhen International Graduate School, Tsinghua University}
  \city{Shenzhen}
  \country{China}}
\email{chenxiao24@mails.tsinghua.edu.cn}

\author{Huandong Wang}
\affiliation{%
  \institution{Department of Electronic Engineering, Tsinghua University}
  \city{Beijing}
  \country{China}}
\email{wanghuandong@tsinghua.edu.cn}

\author{Baining Zhao}
\affiliation{%
 \institution{Tsinghua Shenzhen International Graduate School, Tsinghua University}
 \institution{Peng Cheng Labotory}
 \city{Shenzhen}
 \country{China}}
\email{zbn22@mails.tsinghua.edu.cn}

\author{Han Li}
\affiliation{%
  \institution{Tsinghua Shenzhen International Graduate School, Tsinghua University}
  \city{Beijing}
  \country{China}}
\email{h-li23@mails.tsinghua.edu.cn}

\author{Xinlei Chen}
\authornote{Corresponding author}
\affiliation{%
 \institution{Tsinghua Shenzhen International Graduate School, Tsinghua University}
 \institution{Peng Cheng Labotory}
  \city{Shenzhen}
  \country{China}}
\email{chen.xinlei@sz.tsinghua.edu.cn}



\begin{abstract}
Images shared online strongly influence emotions and public well-being. Understanding the emotions an image \emph{elicits} is therefore vital for fostering healthier and more sustainable digital communities, especially during public crises. We study \textit{Visual Emotion Elicitation} (\VEE), predicting the \emph{set} of emotions that an image evokes in viewers. We introduce \textbf{\VAEER}, an interpretable multi-label \VEE framework that combines attention-inspired cue extraction with knowledge-grounded reasoning. \VAEER isolates salient visual foci and contextual signals, aligns them with structured affective knowledge, and performs per-emotion inference to yield transparent, emotion-specific rationales. Across three heterogeneous benchmarks—including social imagery and disaster-related photos—\VAEER achieves state-of-the-art results with up to \textbf{19\%} per-emotion improvements and a \textbf{12.3\%} average gain over strong CNN and VLM baselines. Our findings highlight interpretable multi-label emotion elicitation as a scalable foundation for responsible visual media analysis and emotionally sustainable online ecosystems.
\end{abstract}



\maketitle
\section{Introduction}
\label{sec:intro}
With visual content becoming central to online communication, the emotions conveyed and evoked by web images play a critical role in shaping collective well-being, public discourse, and the dynamics of online communities. On one hand, images shared on social platforms can amplify empathy, foster prosocial behavior, and support social connection~\cite{yang2023emoset}. On the other hand, they can escalate anxiety, spread distress, and reinforce harmful narratives when circulated at scale~\cite{wang2023image}. Understanding these emotional signals is therefore essential for building healthier and more sustainable information ecosystems, aligning with the United Nations Sustainable Development Goals, particularly \textbf{SDG~3} (Good Health and Well-Being), \textbf{SDG~11} (Sustainable Cities and Communities), and \textbf{SDG~13} (Climate Action). 

We consider the problem of \textit{Visual Emotion Elicitation} (\VEE)~\cite{peng2015mixed,wang2023image}: \textit{What emotions does an image evoke in its viewers?}. While \VEE has traditionally been studied in affective computing and creative AI, its societal relevance is increasingly evident. In crisis informatics and public health, emotion-aware analysis enables real-time monitoring of community stress, early detection of collective distress, and more empathetic disaster communication~\cite{kuvsen2020emotional,halse2018emotional}. For sustainable information ecosystems, it supports moderation of highly distressing or manipulative content, helping reduce psychological harm while preserving situational awareness. During natural disasters, emotion-sensitive interpretation of disaster imagery informs the design of responsible media that raises awareness without inducing unnecessary fear.

Although emotional responses may appear subjective, cognitive and neuroscientific evidence reveal that image-evoked affect follows consistent patterns~\cite{kahneman2011thinking, tooby2008evolutionary, kragel2019emotion, yang2023emoset, hassan2020visual}. \textit{Dual-process theory} shows that affective appraisal is rapid and automatic (\textit{System~1}), often bypassing deliberate cognition. \textit{Biological Preparedness} explains why humans react predictably to ecologically salient cues such as threat, destruction, or human suffering. \textit{Visual emotion schemas} further demonstrate that low-level perceptual signals can directly trigger affective states. Furthermore, empirical evidence of cross-cultural emotion annotation studies confirms that image-elicited emotions exhibit cross-population regularity. These findings support a central premise: \VEE offers a principled pathway to computationally model affective impact at the population scale, with meaningful implications for community well-being, especially during public crises.

Early computational approaches explored \VEE through CNNs focusing on facial expressions~\cite{yu2015image}, body language~\cite{zhu2017dependency}, and artistic attributes~\cite{jing2023styleedl, peng2015mixed}. Subsequent models incorporated contextual cues~\cite{lee2019context, huang2021emotion, yang2022emotion}, but often struggled with semantic grounding and ambiguity. Recent vision-language models (VLMs)~\cite{achiam2023gpt, yang2024emollm, xie2024emovit} introduce powerful general reasoning abilities, yet remain limited in \textit{emotionally complex, multi-label} settings typical of disaster scenes, where multiple interacting cues collectively shape affective appraisal.

We identify three major challenges that hinder socially responsible and practically deployable \VEE systems.  
First, \VEE is inherently \emph{multi-label}: social media images commonly evoke \emph{multiple} emotions (e.g., fear, sadness, empathy) simultaneously, yet many systems collapse predictions to a single dominant label, obscuring important affective nuance.  
Second, there is a gap in \emph{emotional grounding} of visual elements. VLM embeddings align visual cues to objects and entities well, but map poorly to \emph{abstract} affective concepts, limiting robustness under ambiguous or high-stress scenarios~\cite{xie2024emovit}.
Third, emotional drivers in images are highly complex and entangled. Prior work suggests that emotional appraisal arises from interacting \textit{Visual Emotion Cues} (\VECs), including human expressions~\cite{wild2001emotions}, environmental conditions~\cite{zhang2024visual}, lighting~\cite{wei2020learning}, and color~\cite{machajdik2010affective}. These \VECs are difficult to identify, structure, and relate in an interpretable way.  

\begin{table*}[]
    \centering
    \caption{Visual attention-inspired categorization of specific \VECs reported in prior literature.}
    \label{tab: VECs}
    \begin{tabularx}{\textwidth}{|l|X|}
    \hline
        Visual Attention-Inspired Category & Visual Emotion Cues   \\ \hline
        Visual Focus~\cite{fan2022emotional} & Human~\cite{lu2024gpt} (Action~\cite{yang2023emoset, yang2024emollm}, Facial Expression~\cite{yang2023emoset, zhang2024visual}, Body Language~\cite{zhang2024visual, mittal2020emoticon}), Animal~\cite{lu2024gpt}, Object~\cite{yang2023emoset}, Interactions~\cite{mittal2020emoticon} \\ \hline
        Visual Context~\cite{mittal2020emoticon} & Environment~\cite{zhang2024visual}, Scene, Composition~\cite{peng2015mixed}, Style~\cite{wang2023unlocking} (Texture~\cite{machajdik2010affective}, Color~\cite{machajdik2010affective, yang2023emoset, peng2015mixed}, Shape~\cite{joshi2011aesthetics}, Edge~\cite{peng2015mixed}, Symmetry~\cite{peng2015mixed}, Saturation~\cite{machajdik2010affective}, Brightness~\cite{yang2023emoset}, Contrast~\cite{machajdik2010affective}, Homogeneity~\cite{machajdik2010affective, peng2015mixed}, Hue~\cite{machajdik2010affective}) \\ \hline
    \end{tabularx}
\end{table*}

To address these challenges, we propose \textbf{\VAEER}, a visual attention-inspired emotion elicitation reasoning framework for multi-label \VEE. Guided by theories of selective attention and contextual modulation~\cite{fan2022emotional,mittal2020emoticon}, \VAEER explicitly disentangles emotional drivers into \textbf{Visual Focus (VF)}, \textbf{Visual Context (VC)}, and their interplay. Technically, \VAEER introduces: (i) a \emph{Visual Attention Masking} module that uses a VLM to localize and describe salient foci, context, and their relationships, (ii) a \emph{Multi-Modal Emotion Retrieval-Augmented Generation} module that aligns these cues with a structured affective knowledge graph for emotion grounding, and (iii) a \emph{Per-Emotion Arousal Reasoning} scheme that performs emotion-specific reasoning over both attention-derived cues and retrieved knowledge to produce transparent multi-label predictions.

Our contributions are summarized as follows:
\begin{packed_itemize}
    \item Grounded in cognitive theories of human appraisal, we design a \emph{Visual Attention Masking} module that decomposes an image into \textbf{Visual Focus}, \textbf{Foci Interactions}, \textbf{Visual Context}, and \textbf{Foci-Context Relationships}. This yields structured, interpretable cues that mirror human attention and scene understanding.
    \item We integrate VLM representations with a structured affective knowledge graph via a \emph{Multi-Modal Emotion Retrieval-Augmented Generation} module. By jointly leveraging textual and visual embeddings, MME-RAG retrieves psychologically informed emotion concepts that provide robust semantic grounding for abstract emotions across diverse real-world images.
    \item We introduce a \emph{Per-Emotion Arousal Reasoning} module that fuses attention-derived cues and retrieved knowledge to reason independently about each emotion category. It produces multi-label predictions with traceable rationales and identifies reinforcing or contradictory evidence among different signals.
    \item We evaluate \VAEER on three representative social media benchmarks under diverse themes, including social imagery, human-centered scenes, and disaster-related photographs. \VAEER achieves state-of-the-art performance with up to \textbf{19\%} per-emotion gains and a \textbf{12.3\%} average improvement over strong CNN and VLM baselines, highlighting its potential as a scalable foundation for emotion-aware visual media analysis toward healthier, more emotionally sustainable online communities.
\end{packed_itemize}

\section{Preliminaries}
\label{sec: preliminaries}

\subsection{Problem Definition}
\noindent\textbf{Multi-Label Visual Emotion Elicitation.}  
Let $\mathbf{G}$ denote the space of images and $\mathbf{A}=\{e_1,\dots,e_n\}$ the emotion set. For $g\in\mathbf{G}$, the goal is to estimate a subset $\mathbf{E}\subseteq\mathbf{A}$ of evoked emotions:
\begin{equation}
\label{eqn: VEA}
\mathbf{E} = f(g,\cdot),
\end{equation}
where $f$ is the prediction function and $\cdot$ denotes optional auxiliary inputs (e.g., metadata, annotations). Unlike single-label classification, \VEE requires reasoning over multiple coexisting emotions, which often emerge from complex interactions among different visual cues. Traditionally, researchers address the problem of \VEE with proposed visual emotion cues as shown in the second column of Table~\ref{tab: VECs}.

\subsection{Cognitive and Neuroscience Foundations}
Understanding what images elicit systematic emotional responses requires bridging cognitive psychology and neuroscience. Decades of research identify two complementary mechanisms in human vision: \textit{selective focus} on salient elements and \textit{contextual surroundings} integrations. Together, these mechanisms suggest that computational \VEE should explicitly account for both \textit{visual foci} and \textit{visual context}. Below, we expand on foundational theories.

\vspace{0.5em}\noindent
\textbf{Selective Visual Attention.}  
Humans cannot process all the details of a scene simultaneously. Visual attention theory shows that perception prioritizes certain features or regions that stand out due to salience~\cite{treisman1980feature, itti2001computational}. In affective vision, this becomes \textit{emotional selective attention}, where emotionally charged stimuli (e.g., a cute animal, a fearful face, or a weapon) dominate appraisal~\cite{fan2022emotional}. Neuroscientific studies on \textit{visual emotion schemas}~\cite{kragel2019emotion} confirm that such salient stimuli can trigger fast, reflex-like responses in the visual cortex, bypassing deliberate reasoning. This explains why localized visual foci often act as anchors of primary affective response. When multiple foci coexist, attentional competition and integration~\cite{desimone1998visual, schupp2003emotional} determine whether emotions reinforce, contrast, or compete. 

\vspace{0.5em}\noindent
\textbf{Visual Context Modulation.}  
Theories such as Frege’s \textit{context principle}~\cite{mittal2020emoticon} and \textit{contextual congruence theory}~\cite{xu2017facial} emphasize that meaning emerges only in relation to surroundings. Neuroscience further shows that global scene information, such as lighting, spatial composition, and environmental setting, modulates interpretation at early perceptual stages~\cite{bar2004visual, oliva2007role}. The contextual scene sets the backbone to address \VEE.

\vspace{0.5em}\noindent
\textbf{Feature Integration.}  
Treisman’s \textit{feature integration theory}~\cite{treisman1980feature} demonstrates that perception arises from binding multiple features (color, shape, motion) into coherent objects. Affective appraisal operates analogously: emotions emerge from the joint processing of local foci and global context. For example, an image of a smiling face in a festive environment would evoke joy, while the same face in a devastated war setting might surprise people. Thus, the emotional weight of foci is stabilized or altered by their embedding context. This hierarchical integration mirrors how the visual cortex and higher-order brain areas combine signals to construct a unified affective interplay. In \VEE, this suggests that both foci and contextual cues, along with their interactions, must be modeled rather than treated independently.

\vspace{0.5em}
The above work shows that human emotional responses to an image are intrinsically influenced by (i) the objects that capture visual attention, (ii) the contextual scene in which they are embedded, and (iii) the interplay among them.

\subsection{Visual Attention Concepts}
Building on these theoretical foundations, we consolidate prior work on \textit{Visual Emotion Cues} (\VECs) into a visual attention-inspired taxonomy, summarized in Table~\ref{tab: VECs}. We formalize the following definitions:

\vspace{0.5em}
\noindent\textbf{Visual Emotion Cue (VEC).}  
A perceptual feature, element, or attribute within an image that contributes to eliciting emotional responses in viewers. Typically expressed as concise natural language phrases, \VECs serve as semantic units for grounding in \VEE. Literature identifies diverse \VECs, including facial expressions~\cite{wild2001emotions}, lighting~\cite{wei2020learning}, and color~\cite{machajdik2010affective}, among others as shown in~\ref{tab: VECs}.

\vspace{0.5em}
\noindent\textbf{Visual Focus (VF).}  
One or more salient regions, objects, or elements that preferentially capture perceptual attention. Grounded in visual attention theory~\cite{treisman1980feature, itti2001computational}, Visual Foci dominate perceptual resources and often anchor emotional appraisal.

\vspace{0.5em}
\noindent\textbf{Foci Interactions (FI).}  
The second-order affective dynamics that emerge when multiple foci coexist. Relationships such as reinforcement, contrast, or competition shape collective appraisal. Attentional competition~\cite{desimone1998visual} and affective integration~\cite{schupp2003emotional} suggest that FI can intensify, balance, or destabilize emotional meaning.

\vspace{0.5em}
\noindent\textbf{Visual Context (VC).}  
The surrounding background, global scene, and structural layout that situate focal elements. Scene perception theory~\cite{bar2004visual} and contextual modulation theory~\cite{oliva2007role} show that VC provides semantic priors that shape interpretation and co-determine affective appraisal.

\vspace{0.5em}
\noindent\textbf{Foci–Context Relationships (FCR).}  
The bidirectional interplay between salient foci and their embedding scene. Congruent contexts can amplify focal emotional signals, while incongruent contexts attenuate or invert them. Conversely, strong foci may reshape how the context is emotionally construed. FCR thus captures hierarchical alignment between local and global affective information.

\vspace{0.5em}
\noindent In summary, these clarifications unify prior empirical findings into a visual attention-inspired taxonomy of \VECs. This taxonomy provides the theoretical backbone for our framework, which explicitly models both focus–focus and focus–context dynamics in multi-label \VEE.

\begin{figure*}[t]
    \centering    
    \includegraphics[width=\linewidth]{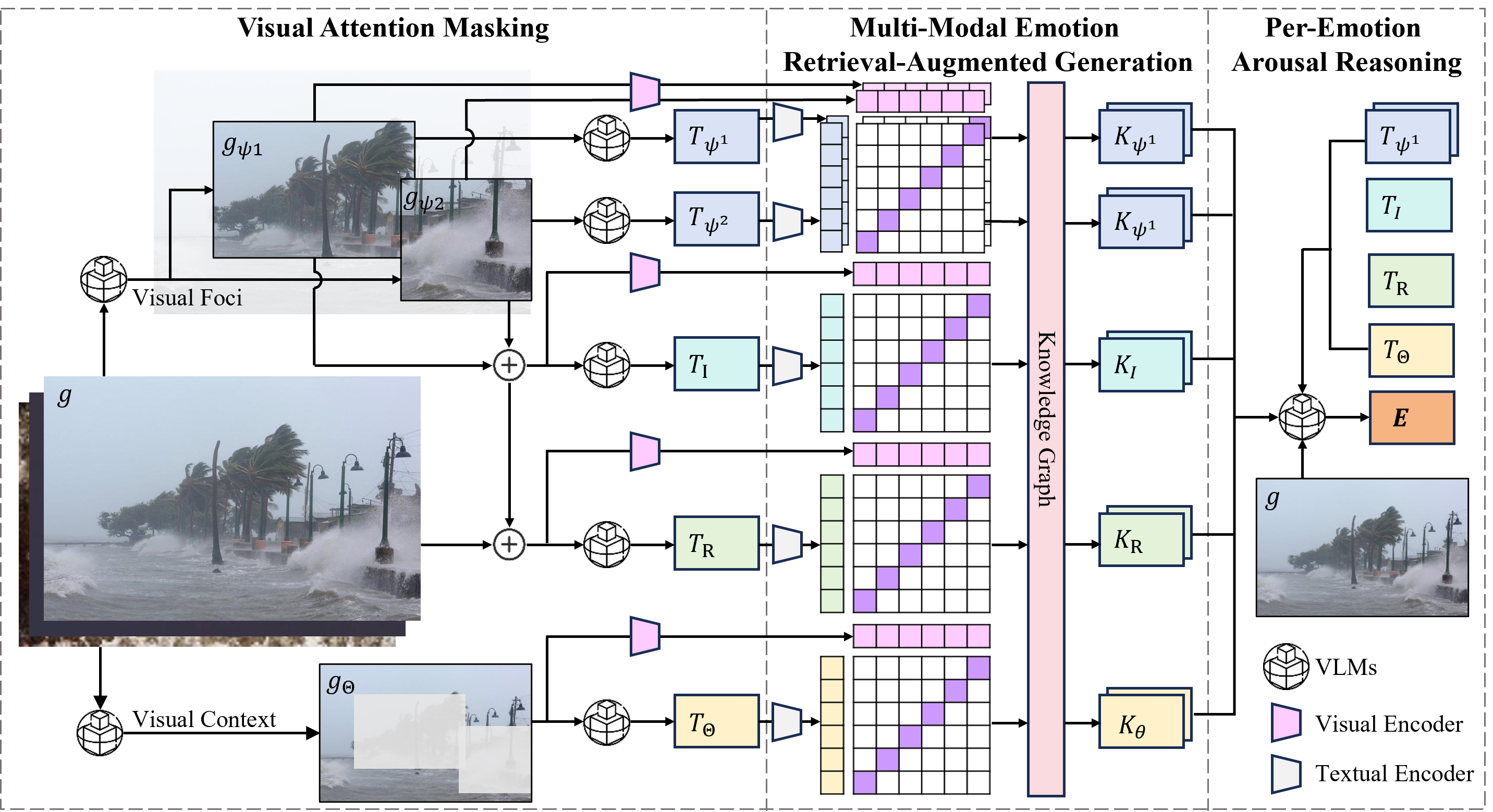}
    \caption{Overview of \VAEER. A VLM extracts visual foci ($\psi^1,\psi^2,\dots$), visual context ($\Theta$) through masking and produces emotion-specific annotations $T_\psi, T_{\mathbf{\Psi}}, T_\Theta, T_{(\Theta,\mathbf{\Psi})}$. Then, the textual and visual embeddings are aligned for emotional grounding through an affective knowledge graph to retrieve top-$k$ emotion concepts $K_\psi, K_{\mathbf{\Psi}}, K_\Theta, K_{(\Theta,\mathbf{\Psi})}$. The model integrates retrieved concepts with the original image to infer the multi-label set $\mathbf{E}$.}
    \label{fig: framework}
\end{figure*}

\section{Methodology}
\label{sec:methodology}

We present \textbf{\VAEER}, a three-stage framework that unifies attention-inspired extraction, knowledge-grounded retrieval, and per-emotion arousal reasoning. Guided by the foundations in Section~\ref{sec:preliminaries}, \VAEER explicitly models visual focus, foci interactions, visual context, and foci--context relationships to produce interpretable multi-label predictions.

\subsection{Overall Framework}

The \VAEER architecture fuses vision--language reasoning with structured emotion knowledge to perform multi-label \VEE based on cognitive and neuroscience findings on the relationship between visual attention and emotion. As illustrated in Fig.~\ref{fig: framework}, the framework consists of three core modules:

\begin{packed_itemize}
    \item \textbf{Visual Attention Masking (VAM):} Mimics human selective attention to identify salient visual foci, their interactions, surrounding context, and foci--context relationships. VAM processes an image into structured textual annotations that capture such cues.
    \item \textbf{Multi-Modal Emotion Retrieval-Augmented Generation (MME-RAG):} Grounds these annotations in a structured affective knowledge graph. MME-RAG aligns visual and textual embeddings and retrieves semantically relevant emotion concepts for each type of cue.
    \item \textbf{Per-Emotion Arousal Reasoning (PAR):} Performs emotion-specific analysis by integrating attention-derived annotations, retrieved concepts, and the original image to infer the final set of evoked emotions and their activation strength.
\end{packed_itemize}

\subsection{Visual Attention Masking (VAM)}
\label{sec:vam}

Guided by the taxonomy in Section~\ref{sec: preliminaries}, the VAM module emulates human appraisal by (i) identifying salient visual foci, (ii) interpreting their interactions, (iii) examining the surrounding visual context, and (iv) reasoning about foci--context relationships.

\vspace{0.5em}
\noindent\textbf{Extraction of Visual Foci.}
Given an input image $g\in\mathbf{G}$, a VLM is prompted to identify one or more regions that capture selective visual attention:
\begin{equation}
\label{eq:vam_foci}
\mathbf{\Psi} = \text{VLM}(g, i_{\mathbf{\Psi}}),
\end{equation}
where $\mathbf{\Psi}=\{\psi^1,\psi^2,\dots\}$ denotes the set of extracted visual foci.  
Each region $\psi\in\mathbf{\Psi}$ yields a crop $g_{\psi}$ for further analysis.

\vspace{0.5em}
\noindent\textbf{Annotations of Visual Foci and Their Interactions.}
Every focus crop is passed to the VLM to produce a concise semantic description with respect to instruction $i_{\psi}$:
\begin{equation}
T_{\psi} = \text{VLM}(g_{\psi}, i_{\psi}), \quad \forall \psi\in\mathbf{\Psi}.
\end{equation}
We denote the collection of VF annotations as 
$\mathcal{C}(T_{\psi},\mathbf{\Psi})=\{T_{\psi} : \psi \in \mathbf{\Psi}\}$.  
To capture FI, the VLM additionally describes their collective configuration and interaction dynamics:
\begin{equation}
T_{I} = \text{VLM}\big(\mathcal{C}(g,\mathbf{\Psi}), i_{I}\big),
\end{equation}
where $\mathcal{C}(g,\mathbf{\Psi})=\{g_{\psi} : \psi \in \mathbf{\Psi}\}$ collects all focus crops.

\vspace{0.5em}
\noindent\textbf{Visual Context Masking and Annotation.}
To capture global contextual information, we mask out all foci from the original image:
\[
g_{\Theta} = \text{MaskOut}(g, \mathbf{\Psi}),
\]
producing an image that contains only the visual context. The VLM generates its corresponding description:
\begin{equation}
T_{\Theta} = \text{VLM}(g_{\Theta}, i_{\Theta}),
\end{equation}
which summarizes the background scene, environment, and global layout.

\vspace{0.5em}
\noindent\textbf{Foci--Context Relationships.}
Affective appraisal also depends on how salient elements relate to their surroundings.  
We therefore instruct the VLM to delineate the relationship between the foci and the contextual scene:
\begin{equation}
T_R = \text{VLM}\big(\mathcal{C}(g,\mathbf{\Psi}), g_{\Theta}, i_{R}\big),
\end{equation}
capturing \textit{Foci--Context Relationships} (FCR) such as congruence, contrast, or tension between local and global cues.

\vspace{0.5em}
The VAM module compiles all generated annotations into:
\begin{equation}
\mathbf{T} = \big\{\mathcal{C}(T_{\psi},\mathbf{\Psi}),\; T_{I},\; T_{\Theta},\; T_{R}\big\},
\end{equation}
representing (i) individual VF annotations, (ii) focus--focus interactions,  
(iii) global context cues, and (iv) foci--context relationships.  
These interpretable components serve as structured inputs for the subsequent emotion grounding and reasoning stages.

\subsection{Multi-Modal Emotion Retrieval-Augmented Generation (MME-RAG)}
\label{sec:rag}

While VAM provides structured and interpretable semantic cues, VLMs alone do not possess explicit grounding in psychologically validated emotion knowledge. To bridge this gap, we introduce a multi-modal emotion retrieval-augmented generation module that aligns extracted \VECs with a structured emotion knowledge graph. This stage contextualizes local and global cues within a broader affective conceptual space.

\vspace{0.5em}
\noindent\textbf{Emotion Knowledge Graph.}
The knowledge graph consists of nodes representing emotion concepts (e.g., \textit{fear}, \textit{awe}, \textit{calmness}), related appraisals, and prototypical situations, with edges encoding semantic or psychological relations (e.g., co-occurrence, causality, similarity). Each node is associated with a text embedding $\gamma$ obtained from the same encoder as the VLM text branch.

\vspace{0.5em}
\noindent\textbf{Embedding Alignment.}
From the VAM outputs $\mathbf{T}$, we obtain textual embeddings $\mathcal{T}$ by encoding each annotation and aggregating them (e.g., by mean pooling or concatenation). From the original image $g$, we obtain visual embeddings $\mathcal{V}$ from the VLM image encoder. For a graph concept embedding $\gamma$, we compute similarity scores with both modalities:
\begin{equation}
\label{eqn: sim}
\lambda_t(\gamma)=\frac{\mathcal{T}^\top \gamma}{\|\mathcal{T}\|\,\|\gamma\|},\qquad
\lambda_v(\gamma)=\frac{\mathcal{V}^\top \gamma}{\|\mathcal{V}\|\,\|\gamma\|}.
\end{equation}

\vspace{0.5em}
\noindent\textbf{Cognitively Inspired Multi-Modal Fusion.}
Because human affective appraisal integrates textual and visual channels, we fuse the two similarity signals using a tunable weight $\alpha\in[0,1]$:
\begin{equation}
\label{eqn: weight}
\lambda_m(\gamma) = \alpha\,\lambda_t(\gamma) + (1-\alpha)\,\lambda_v(\gamma).
\end{equation}
This fusion mechanism allows the model to adaptively emphasize linguistic or perceptual cues depending on the reliability of each modality, consistent with theories of multi-modal integration in cognitive science.

\vspace{0.5em}
\noindent\textbf{Emotion Concept Retrieval.}
We retrieve the top-$k$ affective concepts most aligned with the fused embedding signal:
\begin{equation}
\label{eqn: RAG}
\mathbf{K} = \text{TopK}(\lambda_m, k).
\end{equation}
For downstream interpretability, retrieved concepts may be grouped by their originating cues (e.g., per-focus, interaction-level, context-level, relation-level), denoted generically as 
\[
\mathcal{C}(K_{\psi},\mathbf{\Psi}),\; K_{I},\; K_{\Theta},\; K_{R}.
\]
These retrieved concepts operationalize how affective meaning is grounded in established psychological knowledge.

\subsection{Per-Emotion Arousal Reasoning (PAR)}
\label{sec:ear}

Given the structured cues $\mathbf{T}$ extracted in VAM and the grounded emotional concepts $\mathbf{K}$ retrieved in MME-RAG, the final stage performs emotion-specific reasoning to determine which emotions are sufficiently activated by the image. Since different cues may reinforce, compete, or contradict one another, we adopt a per-emotion schema that reasons independently over each candidate emotion.

\vspace{0.5em}
\noindent\textbf{Emotion-Aware Integrative Reasoning.}
For each emotion category $e\in\mathbf{A}$, the VLM is prompted with a specialized per-emotion instruction $i_e$ that integrates (i) visual foci, (ii) interactions between foci, (iii) contextual cues, (iv) foci--context relationships, and (v) retrieved affective concepts. This design ensures that the reasoning process accounts for alignment and contradiction across multiple evidence sources. In particular, consensus among VF/FI/VC/FCR cues is rewarded, whereas systematic disagreement triggers penalization, echoing psychological observations of cue integration in human affective judgment.

Formally, the predicted emotion set is produced as:
\begin{equation} 
\label{eqn: alpha}
\mathbf{E} = \text{VLM}(\mathbf{T}, \mathbf{K}, g, i_p),
\end{equation}
where the VLM infers arousal levels and determines whether each emotion should be included in the final label set. The output preserves traceability: each decision is supported by interpretable evidence from both the visual attention pathway and the knowledge-grounding pathway.

\vspace{0.5em}
\noindent\textbf{Interpretability and Theoretical Alignment.}
This stage unifies local (VF, FI) and global (VC, FCR) evidence with structured knowledge, enabling the model to reason about emotionally relevant contradictions. For example, visual foci may express danger while the context suggests calmness, or conversely, multiple cues may jointly signal intense distress. The modular design ensures that emotions are predicted not only by pattern recognition but by cognitively motivated appraisal-like reasoning.

\vspace{0.5em}
\noindent In summary, Stage~I (\textbf{VAM}) disentangles affective cues through attention-driven decomposition, Stage~II (\textbf{MME-RAG}) grounds these cues using a psychologically informed emotion knowledge graph, and Stage~III (\textbf{PAR}) synthesizes both sources into a coherent multi-label prediction $\mathbf{E}$ with transparent and semantically meaningful rationales. The full rundown is attached in Appendix~\ref {appendix: instructions}.

\begin{table*}[]
\caption{The performance of different \VEE methods evaluated per emotion across three datasets. Bold indicates the highest accuracy and F1 score, and underlining denotes the second-best results.}
\label{tab: experiment}
\resizebox{\textwidth}{!}{%
\begin{tabular}{ccccccccccccccccc}
\hline
\multicolumn{1}{c|}{{}}                                  & \multicolumn{2}{c|}{{\textbf{Sadness}}}                                       & \multicolumn{2}{c|}{{\textbf{Joy}}}                                           & \multicolumn{2}{c|}{{\textbf{Fear}}}                                          & \multicolumn{2}{c|}{{\textbf{Disgust}}}                                       & \multicolumn{2}{c|}{{\textbf{Anger}}}                                         & \multicolumn{2}{c|}{{\textbf{Surprise}}}                                      & \multicolumn{2}{c|}{{\textbf{Neutral}}}                                       & \multicolumn{2}{c}{{\textbf{Overall}}}                                       \\ 
\multicolumn{1}{c|}{\multirow{-2}{*}{{\textbf{Method}}}} & {Acc}                       & \multicolumn{1}{c|}{{F1}}   & {Acc}                       & \multicolumn{1}{c|}{{F1}}   & {Acc}                       & \multicolumn{1}{c|}{{F1}}   & {Acc}                       & \multicolumn{1}{c|}{{F1}}   & {Acc}                       & \multicolumn{1}{c|}{{F1}}   & {Acc}                       & \multicolumn{1}{c|}{{F1}}   & {Acc}                       & \multicolumn{1}{c|}{{F1}}   & {Acc}                       & {Avg-F1}                       \\ \hline

\multicolumn{17}{c}{{\textbf{Emotion6}}}                                                                                                                                                                                                                                                                               \\ \hline
\multicolumn{1}{c|}{{Emotic}}                            & {0.59}                     & \multicolumn{1}{c|}{{0.50}} & {0.52}                     & \multicolumn{1}{c|}{{0.57}} & {0.55}                     & \multicolumn{1}{c|}{{0.47}} & {0.46}                     & \multicolumn{1}{c|}{{0.19}} & {0.62}                     & \multicolumn{1}{c|}{{0.38}} & {0.50}                     & \multicolumn{1}{c|}{{0.30}} & {0.56}                     & \multicolumn{1}{c|}{{0.40}} & {0.54}                     & {0.40}                     \\
\multicolumn{1}{c|}{{StyleEDL}}                          & {0.61}                     & \multicolumn{1}{c|}{{0.55}} & {0.68}                     & \multicolumn{1}{c|}{{0.42}} & {0.57}                     & \multicolumn{1}{c|}{{0.61}} & {0.53}                     & \multicolumn{1}{c|}{{0.28}} & {0.62}                     & \multicolumn{1}{c|}{{0.35}} & {0.58}                     & \multicolumn{1}{c|}{{0.47}} & {0.56}                         & \multicolumn{1}{c|}{{0.40}}     & {0.60}                     & {0.45}                     \\
\multicolumn{1}{c|}{{EmotionNet}}                          & {0.70}                     & \multicolumn{1}{c|}{{0.66}} & {\underline{0.69}}                     & \multicolumn{1}{c|}{{\underline{0.61}}} & {0.63}                     & \multicolumn{1}{c|}{{0.47}} & {0.65}                     & \multicolumn{1}{c|}{{0.48}} & {0.67}                     & \multicolumn{1}{c|}{{0.54}} & {0.51}                     & \multicolumn{1}{c|}{{0.27}} & {0.49}                     & \multicolumn{1}{c|}{{0.25}} & {0.62}                     & {0.65}                     \\

\multicolumn{1}{c|}{{EmoVIT}}                            & {0.68}                     & \multicolumn{1}{c|}{{0.38}} & {0.65}                     & \multicolumn{1}{c|}{{0.29}} & {0.71}                     & \multicolumn{1}{c|}{{0.60}} & {0.59}                     & \multicolumn{1}{c|}{{0.25}} & {0.68}                     & \multicolumn{1}{c|}{{0.32}} & {0.54}                     & \multicolumn{1}{c|}{{0.33}} & {0.60}                         & \multicolumn{1}{c|}{{0.35}}     & {0.64}                     & {0.36}                     \\
\multicolumn{1}{c|}{{EmoCLIP}}                           & {0.75}                     & \multicolumn{1}{c|}{{0.39}} & {0.67}                     & \multicolumn{1}{c|}{{0.44}} & {0.63}                     & \multicolumn{1}{c|}{{\underline{0.65}}} & {0.56}                     & \multicolumn{1}{c|}{{0.37}} & {0.63}                     & \multicolumn{1}{c|}{{0.29}} & {0.61}                     & \multicolumn{1}{c|}{{0.50}} & {0.63}                         & \multicolumn{1}{c|}{{0.42}}     & {0.64}                     & {0.44}                     \\
\multicolumn{1}{c|}{{GPT4V}}                               & {0.78}                         & \multicolumn{1}{c|}{{0.59}}     & {0.54}                         & \multicolumn{1}{c|}{{0.56}}     & {0.32}                         & \multicolumn{1}{c|}{{0.39}}     & {0.65}                         & \multicolumn{1}{c|}{{0.55}}     & {0.17}                         & \multicolumn{1}{c|}{{0.27}}     & {0.51}                         & \multicolumn{1}{c|}{{0.41}}     & {0.55}                         & \multicolumn{1}{c|}{{0.50}}     & {0.50}                     & {0.47}                     \\
\multicolumn{1}{c|}{SenticNet8}                                               & 0.67                                            & \multicolumn{1}{c|}{0.65}                        & 0.62                                            & \multicolumn{1}{c|}{0.58}                        & 0.66                                            & \multicolumn{1}{c|}{0.39}                        & 0.70                                            & \multicolumn{1}{c|}{0.41}                        & 0.75                                            & \multicolumn{1}{c|}{0.26}                        & 0.59                                            & \multicolumn{1}{c|}{0.28}                        & 0.64                                            & \multicolumn{1}{c|}{0.36}                        & 0.65                                            & 0.52                                            \\

\multicolumn{1}{c|}{{\VAEERL}}                         & {\textbf{0.83}}                     & \multicolumn{1}{c|}{{\textbf{0.74}}} & {\underline{0.69}}                     & \multicolumn{1}{c|}{{\textbf{0.63}}} & {\textbf{0.83}}                     & \multicolumn{1}{c|}{{0.61}} & {\textbf{0.89}}                     & \multicolumn{1}{c|}{\underline{0.57}} & {\textbf{0.93}}                     & \multicolumn{1}{c|}{\textbf{0.69}} & {\underline{0.68}}                     & \multicolumn{1}{c|}{\underline{0.57}} & {\underline{0.82}}                     & \multicolumn{1}{c|}{{\underline{0.54}}} & {\textbf{0.81}}                     & {\underline{0.61}}                     \\
\multicolumn{1}{c|}{{\VAEERJ}}                         & \multicolumn{1}{c}{{\underline{0.83}}} & \multicolumn{1}{c|}{{\underline{0.72}}} & \multicolumn{1}{c}{{\textbf{0.73}}} & \multicolumn{1}{c|}{{\underline{0.61}}} & \multicolumn{1}{c}{{\underline{0.71}}} & \multicolumn{1}{c|}{{\textbf{0.70}}} & \multicolumn{1}{c}{{\underline{0.81}}} & \multicolumn{1}{c|}{{\textbf{0.64}}} & \multicolumn{1}{c}{{\underline{0.86}}} & \multicolumn{1}{c|}{{\underline{0.60}}} & \multicolumn{1}{c}{{\textbf{0.72}}} & \multicolumn{1}{c|}{{\textbf{0.59}}} & \multicolumn{1}{c}{{\textbf{0.84}}} & \multicolumn{1}{c|}{{\textbf{0.57}}}     & \multicolumn{1}{c}{{\underline{0.78}}}     & \multicolumn{1}{c}{{\textbf{0.63}}}     \\ \hline
\multicolumn{17}{c}{{\textbf{EmoSet}}}                                                                                                                                                                   \\ \hline
\multicolumn{1}{c|}{{Emotic}}                            & {0.53}                     & \multicolumn{1}{c|}{{0.39}} & {0.50}                     & \multicolumn{1}{c|}{{0.46}} & {0.53}                     & \multicolumn{1}{c|}{{0.38}} & {0.45}                     & \multicolumn{1}{c|}{{0.10}} & {0.54}                     & \multicolumn{1}{c|}{{0.22}} & {0.45}                     & \multicolumn{1}{c|}{{0.26}} & {--}                       & \multicolumn{1}{c|}{{--}}   & {0.50}                     & {0.31}                     \\
\multicolumn{1}{c|}{{StyleEDL}}                          & {0.63}                     & \multicolumn{1}{c|}{{0.47}} & {0.59}                     & \multicolumn{1}{c|}{{0.51}} & {0.55}                     & \multicolumn{1}{c|}{{0.42}} & {0.51}                     & \multicolumn{1}{c|}{{0.25}} & {0.57}                     & \multicolumn{1}{c|}{{0.38}} & {0.49}                     & \multicolumn{1}{c|}{{0.35}} & {--}                       & \multicolumn{1}{c|}{{--}}   & {0.56}                     & {0.40}                     \\
\multicolumn{1}{c|}{{EmotionNet}}                          & {0.72}                     & \multicolumn{1}{c|}{{\underline{0.69}}} & {\underline{0.85}}                     & \multicolumn{1}{c|}{{\textbf{0.85}}} & {0.58}                     & \multicolumn{1}{c|}{{0.30}} & {0.68}                     & \multicolumn{1}{c|}{{0.54}} & {0.70}                     & \multicolumn{1}{c|}{{\textbf{0.62}}} & {0.51}                     & \multicolumn{1}{c|}{{0.21}} & {--}                     & \multicolumn{1}{c|}{{--}} & {0.67}                     & {0.52}                     \\
\multicolumn{1}{c|}{{EmoVIT}}                            & {0.70}                     & \multicolumn{1}{c|}{{0.53}} & {0.73}                     & \multicolumn{1}{c|}{{0.48}} & {0.64}                     & \multicolumn{1}{c|}{{0.57}} & {0.58}                     & \multicolumn{1}{c|}{{0.42}} & {0.65}                     & \multicolumn{1}{c|}{{0.39}} & {0.57}                     & \multicolumn{1}{c|}{{0.45}} & {--}                       & \multicolumn{1}{c|}{{--}}   & {0.65}                     & {0.47}                     \\
\multicolumn{1}{c|}{{EmoCLIP}}                           & {0.68}                     & \multicolumn{1}{c|}{{0.59}} & {0.77}                     & \multicolumn{1}{c|}{{0.56}} & {0.69}                     & \multicolumn{1}{c|}{{0.52}} & {0.62}                     & \multicolumn{1}{c|}{{0.49}} & {0.71}                     & \multicolumn{1}{c|}{{0.47}} & {\underline{0.66}}                     & \multicolumn{1}{c|}{{\textbf{0.53}}} & {--}                       & \multicolumn{1}{c|}{{--}}   & {0.69}                     & {0.53}                     \\
\multicolumn{1}{c|}{{GPT4V}}                               & {0.76}                     & \multicolumn{1}{c|}{{0.70}} & {0.69}                     & \multicolumn{1}{c|}{{0.63}} & {0.33}                     & \multicolumn{1}{c|}{{0.40}} & {0.56}                     & \multicolumn{1}{c|}{{0.53}} & {0.54}                     & \multicolumn{1}{c|}{{0.60}} & {0.38}                     & \multicolumn{1}{c|}{{0.33}} & {--}                       & \multicolumn{1}{c|}{{--}}   & {0.54}                     & {0.53}                     \\
\multicolumn{1}{c|}{SenticNet8}                                               & 0.80                                            & \multicolumn{1}{c|}{0.67}                        & 0.74                                            & \multicolumn{1}{c|}{0.62}                        & 0.68                                            & \multicolumn{1}{c|}{\underline{0.61}}                        & 0.54                                            & \multicolumn{1}{c|}{0.37}                        & 0.76                                            & \multicolumn{1}{c|}{0.28}                        & 0.48                                            & \multicolumn{1}{c|}{0.44}                        & --                                            & \multicolumn{1}{c|}{--}                        & 0.67                                            & 0.50                                            \\
\multicolumn{1}{c|}{{\VAEERL}}                            & {\underline{0.83}}                     & \multicolumn{1}{c|}{{\underline{0.69}}} & {\underline{0.85}}                     & \multicolumn{1}{c|}{{0.61}} & {\underline{0.79}}                     & \multicolumn{1}{c|}{{\textbf{0.62}}} & {\textbf{0.86}}                     & \multicolumn{1}{c|}{{\underline{0.60}}} & {\underline{0.84}}                     & \multicolumn{1}{c|}{{0.55}} & {0.62}                     & \multicolumn{1}{c|}{{0.46}} & {--}                       & \multicolumn{1}{c|}{{--}}   & {\underline{0.80}} & {\underline{0.59}} \\
\multicolumn{1}{c|}{{\VAEERJ}}                            & {\textbf{0.86}}                     & \multicolumn{1}{c|}{{\textbf{0.72}}} & {\textbf{0.88}}                     & \multicolumn{1}{c|}{{\underline{0.65}}} & {\textbf{0.82}}                     & \multicolumn{1}{c|}{{\textbf{0.62}}} & {\underline{0.85}}                     & \multicolumn{1}{c|}{{\textbf{0.68}}} & {\textbf{0.87}}                     & \multicolumn{1}{c|}{{\underline{0.60}}} & {\textbf{0.68}}                     & \multicolumn{1}{c|}{{\underline{0.50}}} & {--}                       & \multicolumn{1}{c|}{{--}}   & {\textbf{0.83}} & {\textbf{0.63}} \\ \hline
\multicolumn{17}{c}{{\textbf{M-Disaster}}}                                                                                        \\ \hline
\multicolumn{1}{c|}{{Emotic}}                            & {0.40}                     & \multicolumn{1}{c|}{{0.44}} & {0.50}                     & \multicolumn{1}{c|}{{0.47}} & {0.51}                     & \multicolumn{1}{c|}{{0.37}} & {0.50}                     & \multicolumn{1}{c|}{{0.19}} & {0.45}                     & \multicolumn{1}{c|}{{0.17}} & {0.49}                     & \multicolumn{1}{c|}{{0.13}} & {0.56}                     & \multicolumn{1}{c|}{{0.42}} & {0.49}                     & {0.27}                     \\
\multicolumn{1}{c|}{{StyleEDL}}                          & {0.76}                     & \multicolumn{1}{c|}{{0.45}} & {0.53}                     & \multicolumn{1}{c|}{{0.35}} & {0.72}                     & \multicolumn{1}{c|}{{0.50}} & {0.47}                     & \multicolumn{1}{c|}{{0.30}} & {0.48}                     & \multicolumn{1}{c|}{{0.32}} & {0.59}                     & \multicolumn{1}{c|}{{0.40}} & {0.60}                     & \multicolumn{1}{c|}{{0.40}} & {0.59}                     & {0.38}                     \\
\multicolumn{1}{c|}{{EmotionNet}}                          & {0.60}                     & \multicolumn{1}{c|}{{0.70}} & {0.58}                     & \multicolumn{1}{c|}{{0.41}} & {0.52}                     & \multicolumn{1}{c|}{{0.22}} & {0.40}                     & \multicolumn{1}{c|}{{0.05}} & {0.51}                     & \multicolumn{1}{c|}{{0.20}} & {0.50}                     & \multicolumn{1}{c|}{{0.23}} & {0.51}                     & \multicolumn{1}{c|}{{0.15}} & {0.53}                     & {0.27}                     \\

\multicolumn{1}{c|}{{EmoVIT}}                            & {0.72}                     & \multicolumn{1}{c|}{{0.42}} & {0.42}                     & \multicolumn{1}{c|}{{0.28}} & {\textbf{0.75}}                     & \multicolumn{1}{c|}{{0.62}} & {0.41}                     & \multicolumn{1}{c|}{{0.27}} & {0.53}                     & \multicolumn{1}{c|}{{0.35}} & {0.34}                     & \multicolumn{1}{c|}{{0.23}} & {0.60}                     & \multicolumn{1}{c|}{{0.44}} & {0.54}                     & {0.37}                     \\
\multicolumn{1}{c|}{{EmoCLIP}}                           & {0.68}                     & \multicolumn{1}{c|}{{0.41}} & {0.55}                     & \multicolumn{1}{c|}{{0.38}} & {0.60}                     & \multicolumn{1}{c|}{{0.60}} & {0.38}                     & \multicolumn{1}{c|}{{0.34}} & {0.25}                     & \multicolumn{1}{c|}{{0.27}} & {0.45}                     & \multicolumn{1}{c|}{{0.46}} & {0.67}                     & \multicolumn{1}{c|}{{0.42}} & {0.51}                     & {0.41}                     \\
\multicolumn{1}{c|}{{GPT4V}}                               & {0.60}                     & \multicolumn{1}{c|}{{0.65}} & {0.67}                     & \multicolumn{1}{c|}{{0.38}} & {0.50}                     & \multicolumn{1}{c|}{{0.36}} & {0.60}                     & \multicolumn{1}{c|}{{0.44}} & {0.59}                     & \multicolumn{1}{c|}{{0.38}} & {0.45}                     & \multicolumn{1}{c|}{{0.23}} & {0.52}                     & \multicolumn{1}{c|}{{0.32}} & {0.51}                     & {0.39}                     \\
\multicolumn{1}{c|}{SenticNet8}                          & 0.74                      & \multicolumn{1}{c|}{0.59}                        & 0.69                                            & \multicolumn{1}{c|}{0.39}                        & 0.65                                            & \multicolumn{1}{c|}{0.61}                        & 0.51                                            & \multicolumn{1}{c|}{0.26}                        & 0.72                                            & \multicolumn{1}{c|}{0.16}                        & 0.48                                            & \multicolumn{1}{c|}{0.33}                        & 0.70                                            & \multicolumn{1}{c|}{0.58}                        & 0.64                                            & 0.40                                            \\
\multicolumn{1}{c|}{{\VAEERL}}                         & {\underline{0.88}}                     & \multicolumn{1}{c|}{{\underline{0.87}}} & {\underline{0.73}}                     & \multicolumn{1}{c|}{{\underline{0.55}}} & {0.68}                     & \multicolumn{1}{c|}{{\underline{0.72}}} & {\underline{0.64}}                     & \multicolumn{1}{c|}{{\underline{0.27}}} & {\underline{0.82}}                     & \multicolumn{1}{c|}{{\underline{0.37}}} & {\underline{0.54}}                     & \multicolumn{1}{c|}{{\underline{0.66}}} & {\underline{0.77}}                     & \multicolumn{1}{c|}{{\underline{0.35}}} & {\underline{0.73}}                     & {\underline{0.54}}                     \\
\multicolumn{1}{c|}{{\VAEERJ}}                         & \multicolumn{1}{c}{{\textbf{0.90}}} & \multicolumn{1}{c|}{{\textbf{0.89}}} & \multicolumn{1}{c}{{\textbf{0.78}}} & \multicolumn{1}{c|}{{\textbf{0.61}}} & \multicolumn{1}{c}{{\underline{0.74}}} & \multicolumn{1}{c|}{{\textbf{0.77}}} & \multicolumn{1}{c}{{\textbf{0.65}}} & \multicolumn{1}{c|}{{\textbf{0.33}}} & \multicolumn{1}{c}{{\textbf{0.84}}} & \multicolumn{1}{c|}{{\textbf{0.39}}} & \multicolumn{1}{c}{{\textbf{0.56}}} & \multicolumn{1}{c|}{{\textbf{0.68}}} & \multicolumn{1}{c}{{\textbf{0.79}}} & \multicolumn{1}{c|}{{\textbf{0.41}}} & \multicolumn{1}{c}{{\textbf{0.75}}} & \multicolumn{1}{c}{{\textbf{0.58}}} \\ \hline
\end{tabular}
}
\end{table*}

\section{Experiment}
\label{sec:experiment}

\subsection{Experimental Settings}

\noindent\textbf{Datasets.}
We evaluate \VAEER on three representative \VEE benchmarks whose labels reflect multiple \emph{evoked} emotions. 
\begin{packed_itemize}
    \item \textbf{Emotion6}~\cite{peng2015mixed} contains 1{,}980 Flickr images, each annotated by more than 15 participants; any emotion with $>67\%$ agreement is treated as evoked, allowing multi-label assignments.  
    \item \textbf{EmoSet-118K}~\cite{yang2023emoset} provides 118{,}102 Flickr images with 10 annotators per image and a 70\% agreement threshold; it also includes \VEC metadata such as brightness, colorfulness, scene type, object categories, facial expressions, and human actions.  
    \item \textbf{M-Disaster}~\cite{hassan2020visual} is the MediaEval Natural Disaster Image Sentiment dataset comprises 3{,}679 Twitter images from natural disasters with 10{,}010 annotations from 2{,}338 participants across 98 countries, plus a \VEC questionnaire identifying dominant influence factors per image.
\end{packed_itemize}


\noindent\textbf{Baselines.}
We compare against strong CNN-, CLIP-, VLM-, and knowledge-graph-based methods, capturing standard deep visual models, instruction-following VLMs, and structured-knowledge approaches.
\begin{packed_itemize} \item \textbf{Emotic}~\cite{kosti2019context}: a CNN model for subject-oriented Visual Emotion Recognition focusing on the emotions of people in the image with context.\item \textbf{StyleEDL}~\cite{jing2023styleedl}: a CNN approach that learns stylistic-aware representations. \item \textbf{EmotionNet}~\cite{wei2020learning}: a noisy-label learning method that exploits weakly supervised web data. \item \textbf{EmoVIT}~\cite{xie2024emovit}: a visual instruction tuning model that fine-tunes InstructBLIP using GPT-generated \VEE-specific training samples. \item \textbf{EmoCLIP}~\cite{foteinopoulou2023emoclip}: an extension of CLIP trained for compound emotion recognition on multiple emotion datasets. \item \textbf{GPT4V}~\cite{achiam2023gpt,hurst2024gpt}: a GPT-4V(o)-based model prompted directly to infer aroused emotions from each image. \item \textbf{SenticNet8}~\cite{cambria2024senticnet}: a multi-modal knowledge-graph framework that enhances emotion understanding and stabilizes LLM outputs via structured reasoning. 
\end{packed_itemize}

\noindent\textbf{Metrics.}
We report per-emotion accuracy and F1. For each emotion $e$, accuracy is computed as
\begin{equation}
\label{acc}
\text{acc}_e = \frac{\text{TP}_e}{\text{Total}_e},
\end{equation}
where $\text{TP}_e$ is the number of correctly predicted positives and $\text{Total}_e$ the number of annotated samples for $e$.  

To address class imbalance in the multi-label setting, we compute F1 on balanced evaluation sets. For each emotion $e$ with $n$ positives, we sample $n$ negatives, repeat this three times, and average:
\begin{equation}
\label{f1-avg}
\text{F1}_e = \frac{1}{3} \sum_{i=1}^{3} 
\frac{2 \cdot \text{TP}_e^{(i)}}{2 \cdot \text{TP}_e^{(i)} + \text{FP}_e^{(i)} + \text{FN}_e^{(i)}}.
\end{equation}
We report per-emotion results and the macro-average over emotions (Avg-F1) in Table~\ref{tab: experiment}.

\noindent\textbf{Implementation Details.}
We instantiate \VAEER with two backbone VLMs: LLaVA-1.6-Vicuna-7B~\cite{liu2024llava} and Janus-Pro-7B~\cite{chen2025janus}, denoted \VAEERL and \VAEERJ. In the MME-RAG stage, we use CLIP~\cite{radford2021learning} as the visual/text encoder and SenticNet8~\cite{cambria2024senticnet} as the affective knowledge graph. The fusion weight $\alpha$ in Eq.~\eqref{eqn: weight} is selected via a two-stage grid search on a held-out validation split: a coarse scan over $[0,1]$ with step 0.1 followed by a finer scan with step 0.05 around the best value. The chosen $\alpha$ is then fixed per backbone across all datasets. Other training and inference hyperparameters follow the original VLM releases unless otherwise noted.

\subsection{Quantitative Evaluation}
\label{sec:quantitative}

Table~\ref{tab: experiment} summarizes per-emotion accuracy and F1 across all datasets. Both \VAEER variants consistently outperform state-of-the-art baselines.

\noindent\textbf{Overall Performance.}  
Averaged across datasets, \VAEERL and \VAEERJ achieve \textbf{0.78} and \textbf{0.79} accuracy, compared to \textbf{0.67} for the strongest baseline, SenticNet8. Avg-F1 shows similar gains, with \VAEERJ improving over SenticNet8 by \textbf{0.11}. Against purely visual CNN baselines (Emotic, StyleEDL), \VAEER improves Avg-F1 by \textbf{0.15–0.25}, demonstrating that decomposing scenes into VF/FI/VC/FCR enhances emotional fidelity beyond object- or style-centric representations. Compared with instruction-tuned or prompt-based VLMs (GPT4V, EmoVIT), \VAEER yields Avg-F1 gains of \textbf{0.10–0.20}, indicating that large VLMs benefit substantially from attention-guided structure and knowledge-grounded alignment rather than direct prompting alone.

\noindent\textbf{Dataset-Level Trends.}  
Across Emotion6, EmoSet, and M-Disaster, \VAEER ranks first in both accuracy and Avg-F1. Gains are most pronounced in high-stress, multi-label settings: on M-Disaster, \VAEERJ improves Avg-F1 by \textbf{0.18} over SenticNet8 and by \textbf{0.21–0.25} over CNN baselines. Even on EmoSet, where EmotionNet is tailored to the dataset, \VAEER achieves \textbf{0.31} accuracy and \textbf{0.11} Avg-F1 improvements, showing that our reasoning-driven approach scales to large heterogeneous collections.

\noindent\textbf{Emotion-Level Patterns.}  
Averaged across datasets, \VAEER yields substantial improvements in accuracy for \emph{sadness} (+0.14), \emph{fear} (+0.12), \emph{anger} (+0.16), and \emph{disgust} (+0.13) over the next-best baseline. These categories heavily depend on interactions between humans, objects, and scene conditions (e.g., damage, danger, degradation), supporting the value of explicitly modeling VF/FI/VC/FCR. For \emph{neutral}, \VAEER avoids over-activating strong emotions, correcting a common failure mode of prompt-only VLMs. \emph{Surprise} remains the most variable class—consistent with its dependence on prior expectations—but \VAEER stays competitive, maintaining higher accuracy and more balanced F1 than most baselines.

Overall, \VAEER improves not only aggregate metrics but also \emph{emotion-specific reliability}, particularly for complex emotions that arise from focus–context interactions.

\subsection{Qualitative Analysis}
\label{sec:qualitative}

To complement quantitative results, we conduct a qualitative case study to examine whether \VAEER’s intermediate reasoning aligns with human appraisal. For each dataset, we randomly sample 100 images and ask three independent annotators to rate VAM, MME-RAG, and PAR outputs. Unless stated otherwise, scores are averaged over annotators on a 1–5 Likert scale, where a score of 5 indicates high agreement. The questionnaire is attached in Appendix~\ref{appendix: questionnaire}.

\noindent\textbf{VAM.}
Annotators judged that visual foci and context were correctly identified in \textbf{97\%} of sampled images. In the remaining 3\%, the model typically included too many minor foci or missed small objects; these did not systematically degrade final predictions. Description quality scores were \textbf{4.65} for VF, \textbf{4.76} for FI, \textbf{4.45} for VC, and \textbf{4.32} for FCR. The mild drop from VF/FI to VC/FCR reflects that people agree more on salient objects and direct interactions than on diffuse context or higher-order relationships, consistent with cognitive findings on visual attention and interpretation.

\noindent\textbf{MME-RAG.}
The relevance of retrieved emotion concepts received an average rating of \textbf{4.64}, indicating that multimodal alignment with SenticNet8 generally produces acceptable, emotionally meaningful concepts that humans find coherent with the scene content. However, it is the lowest agreement score compared to the generated logics. While most concepts are highly relevant and decisive to the final emotion elicitation, a few related yet divergent concepts are identified by the annotators. For example, when the crop of visual focus identified as 'muddy kid' is correlated with "unclean", it is actually due to the disaster in the context. However, the final reasoning step of PAR mitigates such a negative effect by jointly analyzing multiple cues and their effects.

\noindent\textbf{PAR.}
PAR achieved the highest agreement, with an average rating of \textbf{4.87}. Annotators reported that justifications for each predicted emotion were well supported by the identified foci, contextual descriptions, and retrieved concepts, and that the stepwise integration resembled human multi-label emotional appraisal.

\noindent\textbf{Annotator vs.~Dataset Ground Truth.}
The original datasets are human-annotated by larger pools, with a rate of 70\% agreement to be considered as an aroused emotion. In our case study, our annotators disagreed with ground-truth labels in \textbf{8.67\%} of cases. \textbf{88.46\%} disagreements involve \textit{disgust} and \textit{surprise}, emotions known to exhibit high inter-individual variability. \textit{Disgust} is strongly shaped by cultural norms and contamination or moral sensitivity, while \textit{surprise} depends on personal expectations and prior exposure. These discrepancies therefore reflect normal variability in emotional elicitation rather than flaws in the datasets or methods. In many such cases, \VAEER’s predictions aligned with our annotators’ judgments, suggesting that its reasoning is well matched to actual human interpretations.

This qualitative study thus supports that \VAEER’s reasoning is not only interpretable but also cognitively plausible and broadly consistent with human emotional judgment.

\begin{table}[htbp]
\centering
\caption{The overall accuracy of the ablation study removing each component. Bold indicates the highest accuracy.}
\label{tab: ablation}
\begin{tabular}{ccccc}
\toprule
\multirow{2}{*}{Model} & \multirow{2}{*}{Ablation} & \textbf{Emotion6} & \textbf{EmoSet} & \textbf{M-Disaster}  \\
\cmidrule(lr){3-3} \cmidrule(lr){4-4} \cmidrule(lr){5-5}
& & Acc & Acc & Acc \\
\midrule
\multirow{5}{*}{LLaVA-1.6} 
& -VAM & 0.62 & 0.59 & 0.57 \\
& -RAG & 0.58 & 0.60 & 0.57 \\
& -MM & 0.70 & 0.67 & 0.65 \\
& -PAR & 0.77 & 0.75 & 0.71  \\
& \VAEERL & \textbf{0.81} & \textbf{0.80} & \textbf{0.73} \\
\midrule
\multirow{5}{*}{Janus-Pro} 
& -VAM & 0.56 & 0.62 & 0.63 \\
& -RAG & 0.57 & 0.64 & 0.60 \\
& -MM & 0.68 & 0.64 & 0.63 \\
& -PAR & 0.76 & 0.76 & 0.72 \\
& \VAEERJ & \textbf{0.78} & \textbf{0.83} & \textbf{0.75}\\
\bottomrule
\end{tabular}
\end{table}

\subsection{Ablation Studies}
\label{sec:ablation}

To understand the contribution of each component, we perform ablation experiments by removing one module at a time. Table~\ref{tab: ablation} reports overall accuracy on all three datasets for both LLaVA-1.6 and Janus-Pro backbones.

\noindent\textbf{Removing VAM (-VAM).}
Replacing VAM with a single holistic description (no VF/FI/VC/FCR decomposition) causes large drops (e.g., from 0.81 to 0.62 on Emotion6 with LLaVA and from 0.78 to 0.56 with Janus-Pro). This confirms that attention-inspired decomposition is essential: without explicit structure, the model loses fine-grained emotional drivers and behaves similarly to generic VLM prompting.

\noindent\textbf{Removing RAG (-RAG).}
Removing MME-RAG, and thus the affective knowledge grounding, yields accuracies around 0.57–0.60 across backbones and datasets, roughly matching GPT4V-style prompting. In this variant, emotional semantics become unstable and nuanced appraisals collapse, indicating that attention cues alone are insufficient for reliable \VEE without structured affective knowledge.

\noindent\textbf{Removing Multi-Modal Alignment (-MM).}
When we harness retrieval that uses only textual VEC annotations and ignores visual embeddings, performance improves over -VAM and -RAG. However, it remains noticeably below the full model (e.g., 0.70 vs. 0.81 on Emotion6 with LLaVA, 0.68 vs. 0.78 with Janus-Pro). Joint visual–text alignment is therefore critical for grounding retrieved concepts in image-specific evidence, especially for context-heavy emotions such as fear or anger.

\noindent\textbf{Removing PAR (-PAR).}
Removing PAR and applying a single global decision rule instead of per-emotion appraisal lowers accuracy by 3–7 points across datasets (e.g., 0.81 to 0.77 on Emotion6 with LLaVA; 0.83 to 0.76 on EmoSet with Janus-Pro). Treating each emotion as a separate reasoning path thus helps distinguish emotion-specific active vs.~passive states and further resolve conflicting cues in multi-label \VEE.

Across all ablations, the complete \VAEER configuration achieves the highest accuracy for both backbones and on all datasets. VAM and RAG provide the largest gains, while multi-modal alignment and per-emotion appraisal further refine predictions, enabling stable and interpretable reasoning about complex emotional cues.

\subsection{Discussions}
\label{sec:discussion}

\noindent\textbf{Reducing Hallucination Through Structured Reasoning.}
A central advantage of \VAEER is its explicit reasoning pipeline: decomposing visual appraisal into VF, FI, VC, and FCR, then grounding these cues in SenticNet8 before per-emotion decisions. This structure constrains the VLM to reason over concrete, interpretable evidence rather than producing unconstrained end-to-end guesses, thereby mitigating free-form hallucinations often seen in prompt-only VLMs. The high human ratings for intermediate steps (Sec.~\ref{sec:qualitative}) suggest that both the attention decomposition and knowledge grounding anchor emotional inference to semantically coherent cues.

\noindent\textbf{Ethical Implications and Public Good.}
By focusing on evoked emotions rather than only depicted ones, \VAEER is well-positioned for social applications. First, it can help identify content that is likely to trigger strong negative emotions, informing online mental health tools and moderation strategies that reduce unnecessary distress while preserving important information. Second, in crisis scenarios, reliable detection of emotions such as fear, sadness, and confusion can support responders and humanitarian organizations in prioritizing high-impact imagery. Third, grounding decisions in explicit reasoning and knowledge reduces the risk of emotionally misleading outputs, which is essential for countering distressing misinformation during disasters or public emergencies. In all cases, \VAEER’s interpretable rationales facilitate human oversight and accountability.

\noindent\textbf{Limitations and Cultural Considerations.}
Emotional reactions vary across cultures, prior experiences, and individual sensitivities. It is reflected by the 8.67\% disagreement between our annotators and the dataset ground truth, and the 70\% agreement among different annotators for each dataset. Emotions such as \emph{disgust} and \emph{surprise} are particularly sensitive to cultural norms and personal expectations. At the same time, many core reactions to danger, suffering, or environmental threat are broadly shared across populations, especially in natural-disaster contexts, which partially mitigates this concern for our target applications. Future work could incorporate culturally adaptive affective knowledge or user-specific calibration to better capture diversity in emotional elicitation while preserving the benefits of structured, grounded reasoning. Benchmarks of culturally sensitive images annotated by people with diverse cultural backgrounds are essential for such analysis.

\section{Related Works}

\subsection{Visual Affective Computing}
Visual affective computing broadly encompasses two tasks: \textit{Visual Emotion Recognition} (VER) and \textit{Visual Emotion Elicitation} (\VEE).  
VER predicts the emotion \emph{displayed} by humans or animals in an image—typically via facial expressions, gestures, or body language~\cite{leong2023facial, ko2018brief}. In contrast, \VEE predicts the emotion \emph{evoked in viewers} by an image~\cite{gandhi2023multimodal, gill2020review}. While VER is restricted to images with human/animal subjects~\cite{ko2018brief}, \VEE generalizes to all visual content by leveraging contextual, stylistic, and semantic cues including color~\cite{machajdik2010affective, peng2015mixed}, lighting~\cite{wei2020learning}, scene type~\cite{franvek2021viewing}, composition~\cite{peng2015mixed}, aesthetics/style~\cite{jing2023styleedl, xie2024emovit}, storytelling~\cite{horvath2019visual}, cuteness~\cite{buckley2016aww}, and more.  
Although extensive literature identifies numerous \VECs across psychology, sociology, and affective computing~\cite{yang2023emoset, kosti2019context, wild2001emotions}, integrating them into a unified computational framework remains challenging due to their diversity and interactions. 

\subsection{Visual Emotion Elicitation}
Early \VEE approaches were built on CNNs following the success of deep visual features~\cite{krizhevsky2012imagenet}. Methods focused on facial expressions~\cite{yu2015image, zhu2017dependency}, human-centric contexts~\cite{kosti2017emotion, kahou2013combining}, or parallel modeling of foreground and background features~\cite{kosti2017emotion, lee2019context}. Peng et al. and Wei et al. extended CNNs with regression and multimodal fusion~\cite{peng2015mixed, wei2020learning}. More recent work explored stylistic or aesthetic cues using GCNs and hybrid representations~\cite{jing2023styleedl}. However, these models remain limited in two ways: (i) they do not align visual patterns with semantic meaning, and (ii) they struggle to generalize to images where emotional impact arises from interactions between multiple contextual elements.

The introduction of CLIP~\cite{radford2021learning} provided a bridge between visual features and semantic space, enabling stronger grounding of emotional cues. Vision-language models (VLMs)—including LLaVA~\cite{liu2023visual, liu2024llava}, GPT-based models~\cite{yang2023dawn, hurst2024gpt}, and Janus~\cite{ma2024janusflow, chen2025janus}—further enhanced multi-modal reasoning. This led to several emotion-focused extensions: EmoCLIP for facial emotion alignment~\cite{foteinopoulou2023emoclip}, instruction-tuned models such as EmoVIT and EmoLLM for visual or multi-task emotional reasoning~\cite{xing2024emo, xie2024emovit, yang2024emollm}. In parallel, the SenticNet knowledge base~\cite{cambria2010senticnet} evolved into a multimodal resource for grounding emotion concepts and aligning generative models~\cite{cambria2024senticnet}.

Our work follows this line of inquiry by focusing on \textit{viewer-centered emotional impact}, decomposing visual information into interpretable components, and grounding predictions in an affective knowledge graph. By aligning VLM reasoning with human appraisal processes, \VAEER mimics the human visual attention and thereby the human emotional response. It provides a step toward socially responsible multimodal AI—capable of supporting applications in digital well-being, civic communication, and humanitarian response.

\section{Conclusion}
We presented \VAEER, a visual attention–inspired framework for multi-label visual emotion elicitation. \VAEER performs per-emotion reasoning with transparent and reliable explanations. Across three diverse datasets, including challenging disaster-related imagery, \VAEER consistently outperforms state-of-the-art baselines, demonstrating substantial gains in both accuracy and F1.

Beyond technical improvements, our results highlight the value of cognitively grounded and interpretable \VEE for social applications, such as monitoring collective distress and supporting empathetic crisis communication. \VAEER thus offers a principled foundation for responsible visual media analysis in real-world social and crisis contexts.


\bibliographystyle{ACM-Reference-Format}
\bibliography{Kever-sigconf}
\appendix
\section{Appendix}




\subsection{Instruction Rundown of \VAEER}
\label{appendix: instructions}

This section summarizes the full sequence of instructions used in \VAEER.  
The pipeline follows the structure of the main framework as shown in~\ref{fig: framework}, VAM, MME-RAG, and PAR. 

For each sampled image $g$, the VLM receives a series of structured prompts designed to capture visual foci, contextual cues, and emotion-grounded reasoning. The examples below illustrate the format and style of prompts used.

\subsubsection*{1. Visual Attention Masking (VAM)}
\label{appendix: VAM}

\paragraph{1.1 Extracting Visual Foci.}
The VLM is instructed to identify the most salient elements:
\begin{quote}
\small
\texttt{``You are an expert visual analyst. Identify the most important visual focus/foci in this image.  
List each focus as a short descriptive phrase (e.g., ‘weeping baby’, ‘collapsed building’, ‘tall tree’).  
Return a visual focus or a few foci that a human observer would naturally attend to first.''}
\end{quote}
\quad The foci $\Psi = \{\psi^1,\psi^2,\dots\}$ are extracted and the crops $g_\psi$ is obtained based on the textual descriptions.

\paragraph{1.2 Describing Foci Interactions (FI).}
When multiple visual foci exist, we then analyze their interactions objectively.
\begin{quote}
\small
\texttt{``Describe how the visual foci of [$T_{\psi^1}, T_{\psi^2},...$] interact with each other in brief sentences.  
Explain spatial relations and action cues.  
The description should be factual, concise, and grounded in the given images.''}
\end{quote}

\paragraph{1.3 Extracting Visual Context (VC).}
The image of VC is obtained by masking VF. The VLM is instructed to objectively illustrate the context:
\begin{quote}
\small
\texttt{``This is an image with the main characters being masked out. Describe the context of this image in concise phrases, e.g., 'sunny weather', 'bustling streets', 'burning woods'. ''}
\end{quote}

\paragraph{1.4 Describing Foci–Context Relationships (FCR).}
Finally, the VLM connects the foci to their context:
\begin{quote}
\small
\texttt{``Explain how the visual foci [$T_{\psi^1}, T_{\psi^2},...$] relate to the scene context [$T_\Theta$] in brief sentences. 
Describe contrasts, tensions, or dependencies between the focal elements and background. The description should be factual, concise, and grounded in the given images.''}
\end{quote}

\subsubsection*{B. Multi-Modal Emotion Retrieval (MME-RAG)}

From the VAM outputs, we build structured textual cues $\mathbf{C}$ and query the affective knowledge graph.

\paragraph{B1. Building Emotion Queries.}
\begin{quote}
\small
\texttt{``Using the extracted foci, interactions, context, and relationships,  
summarize the emotionally relevant visual cues.  
Return a list of short phrases suitable as semantic queries (e.g., ‘injured people’,  
‘storm damage’, ‘bright festive lights’).''}
\end{quote}

\paragraph{2. MME-RAG.}
For each category of VF, FI, VC, and FCR, the query is obtained by concatenating the obtained phrases from VAM. They are embedded using CLIP and matched to SenticNet8 to retrieve the most relevant concepts

\subsubsection*{3. PAR}

Each emotion $e$ is reasoned independently using the full cue set (VF, FI, VC, FCR, retrieved concepts).

\paragraph{3.1. Emotion-Specific Reasoning Prompt.}
For each emotion label $e$ from sadness, fear, anger, joy, disgust, surprise, neutral:
\begin{quote}
\small
\texttt{``Evaluate whether the following emotion is likely to be evoked in a human viewer: [EMOTION].  
Analyse the information from visual foci [$T_{\psi^1}, T_{\psi^2},...$] and [$K_{\psi^1}, K_{\psi^2},...$], interactions [$T_I$] [$K_I$], contextual cues [$T_\Theta$] [$K_\Theta$] and FCR [$T_R$], [$K_R$]. Analyze how these cues agree or contradict with each other.  
Provide:  
(1) A binary judgment (Yes/No).  
(2) A short justification (2-3 sentences) grounded entirely in the provided cues of the image.''}
\end{quote}

These instructions reflect the full chain-of-thought scaffolding used in \VAEER to encourage structured, interpretable, and non-hallucinatory visual emotion elicitation.

\subsection{Questionnaire for Qualitative Case Study}
\label{appendix: questionnaire}

To evaluate whether \VAEER’s intermediate reasoning aligns with human interpretation, we designed a structured questionnaire administered to three independent annotators. For each dataset, 100 images were randomly sampled. Annotators reviewed the VAM, MME-RAG, and PAR outputs for each image and answered the following items, using a 1–5 Likert scale unless otherwise specified.

\noindent\textbf{Visual Attention Masking (VAM)}
\begin{itemize}
    \item \textbf{VF/VC Correctness (Yes/No):}  
    “Are the detected visual foci and the identified visual context appropriate for this image?”
    \item \textbf{VF Description Quality (1–5):}  
    “How well does the model describe each visual focus?”
    \item \textbf{FI Description Quality (1–5):}  
    “How well does the model describe the interactions among the focal elements?”
    \item \textbf{VC Description Quality (1–5):}  
    “How well does the model describe the global scene context after masking out the foci?”
    \item \textbf{FCR Description Quality (1–5):}  
    “How clearly does the model describe the relationship between focal elements and the surrounding context?”
\end{itemize}

\noindent\textbf{Multi-Modal Emotion Retrieval (MME-RAG)}
\begin{itemize}
    \item \textbf{Concept Relevance (1–5):}  
    “Do the retrieved emotion-related concepts meaningfully match the content and theme of the image?”
\end{itemize}

\noindent\textbf{Per-Emotion Arousal Reasoning (PAR)}
\begin{itemize}
    \item \textbf{Justification Agreement (1–5):}  
    “Does the model’s justification logically support the predicted emotion set?”
\end{itemize}

\noindent\textbf{Agreement With Dataset Ground Truth}
\begin{itemize}
    \item \textbf{Emotion Agreement (Yes/No):}  
    “Do you personally agree with the ground-truth evoked emotions for this image?”  
    Annotators could optionally provide alternative emotion labels with justifications.
\end{itemize}

This questionnaire provides a structured basis for evaluating the interpretability and human alignment of \VAEER’s reasoning process beyond quantitative metrics.









\end{document}